\documentclass{article}
\usepackage{iclr2027_conference,times}

\usepackage{amsmath,amsfonts,bm}

\def\eqref#1{equation~\ref{#1}}

\def\1{\bm{1}}

\DeclareMathAlphabet{\mathsfit}{\encodingdefault}{\sfdefault}{m}{sl}
\SetMathAlphabet{\mathsfit}{bold}{\encodingdefault}{\sfdefault}{bx}{n}

\usepackage{subcaption}
\usepackage[hidelinks]{hyperref}
\hypersetup{
  pdftitle={SkillAdam: Stable and Efficient Skill Evolution for Agents},
  pdfauthor={Gaoyuan Li, Meihao Fan, Yizhe Liu, Shaolei Zhang, Ju Fan, Siyi Wang, Jiaheng Hou, Xudong Weng, Honghan Tian, Zang Li}
}
\usepackage{url}
\usepackage{booktabs}
\usepackage{amsmath}
\usepackage{amssymb}
\usepackage{algorithm}
\usepackage{algorithmic}
\usepackage{graphicx}
\usepackage{float}
\usepackage{enumitem}
\usepackage{xspace}
\newcommand{\sys}{\textsc{SkillAdam}\xspace}

\title{\sys: Stable and Efficient\\Skill Evolution for Agents}

\author{
\textbf{Gaoyuan Li}\textsuperscript{1} \quad \textbf{Meihao Fan}\textsuperscript{1} \quad \textbf{Yizhe Liu}\textsuperscript{1} \quad \textbf{Shaolei Zhang}\textsuperscript{1}\thanks{Corresponding author.} \quad \textbf{Ju Fan}\textsuperscript{1} \\
\textbf{Siyi Wang}\textsuperscript{2} \quad \textbf{Jiaheng Hou}\textsuperscript{2} \quad \textbf{Xudong Weng}\textsuperscript{2} \quad \textbf{Honghan Tian}\textsuperscript{2} \quad \textbf{Zang Li}\textsuperscript{2} \\[0.5ex]
{\small \textsuperscript{1}Renmin University of China \qquad \textsuperscript{2}Tencent} \\
{\small \texttt{\{logey04,fmh1art,liuyizhe2004,zhangshaolei98,fanj\}@ruc.edu.cn}} \\
{\small \texttt{\{skylasywang,marvinhou,steveweng,abeltian,gavinzli\}@tencent.com}}
}

\iclrfinalcopy

\begin{document}

\maketitle
\lhead{}

\begin{abstract}
Agent skills provide a lightweight way to equip frozen language-model agents
with domain knowledge and procedural guidance, yet obtaining high-quality
skills remains costly and difficult to scale. Expert-written skills require
substantial human effort. Recent skill self-evolution methods automate an
iterative loop that uses execution feedback to revise skills, but their
heuristic update strategies often yield unstable optimization and low
{
iteration efficiency. We identify two challenges in realizing stable and
efficient skill self-evolution. \emph{Direction Stability} requires effective
corrections to accumulate rather than be overwritten by iteration-local
feedback. \emph{Update Adaptivity} requires the scope of each revision to
reflect the consistency of recent case-level improvements.
}
We introduce \textbf{\sys}, an Adam-inspired framework for optimizing
discrete and non-differentiable skill documents. As a functional analogue of
Adam's first moment, an optimization memory records identified problems and
the outcomes of prior solution attempts to stabilize the update direction.
As a functional analogue of Adam's second moment, a volatility-driven edit
budget tracks the history-weighted variation of recent case-level improvements
and adaptively controls the update magnitude. Across seven benchmarks that
span short- and long-horizon tasks, \sys achieves state-of-the-art
performance with more stable optimization dynamics. It also obtains stronger
skills with substantially fewer optimization iterations and lower cost than
prior methods.
Code repository: \url{https://github.com/ruc-datalab/SkillAdam}.
\end{abstract}

\section{Introduction}
\label{sec:intro}

Large language model (LLM) agents have emerged as a powerful paradigm for solving complex real-world tasks through multi-step planning and tool use. Representative tasks include web interaction~\citep{wang2025asi}, travel planning~\citep{zhang2026deepplanning}, data preparation~\citep{fan2026deepprep,deng2026dataevolver}, and data discovery and analysis~\citep{zhang2025deepanalyze,zhang2026codabench,liu2026dastudio}. However, LLM agents often struggle in domain-specific scenarios because their general-purpose capabilities cannot satisfy long-tail domain requirements. For example, planning a multi-city trip requires reasoning about visa regulations, transportation dependencies, and airline-specific booking policies. 
To address this challenge, recent agent frameworks introduce \emph{Agent Skills}, defined by Anthropic as modular packages of instructions and supporting resources that equip agents with specialized capabilities~\citep{anthropic2025skills}. 

However, high-quality Skills are both essential and difficult to obtain. SkillsBench shows that carefully curated Skills can substantially improve agent performance~\citep{skillsbench2026}, highlighting the importance of high-quality Skills. In contrast, SkillAxe reports that Skills generated directly by LLMs often remain ineffective without further refinement~\citep{gautam2026skillaxe}, suggesting that automatically constructing high-quality Skills remains a challenging problem. A common solution is to rely on domain experts to manually author Skills based on their expertise, but this process is time-consuming and labor-intensive. Another line of work employs language models to generate or refine Skills using manually designed prompts or heuristic rules. Although these methods reduce the burden of manual Skill authoring, they still rely heavily on human-designed prompts and task-specific heuristics, limiting their ability to generalize across domains.

\begin{figure}[t]
    \centering

    \begin{subfigure}[t]{0.51\linewidth}
        \centering
        \includegraphics[width=\linewidth]{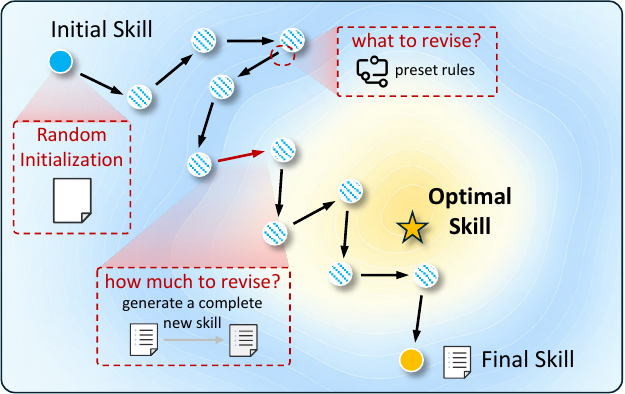}
        \caption{ Existing SGD-like methods.}
        \label{fig:sgd_methods_concept}
    \end{subfigure}
    \hfill
    \begin{subfigure}[t]{0.471\linewidth}
        \centering
        \includegraphics[width=\linewidth]{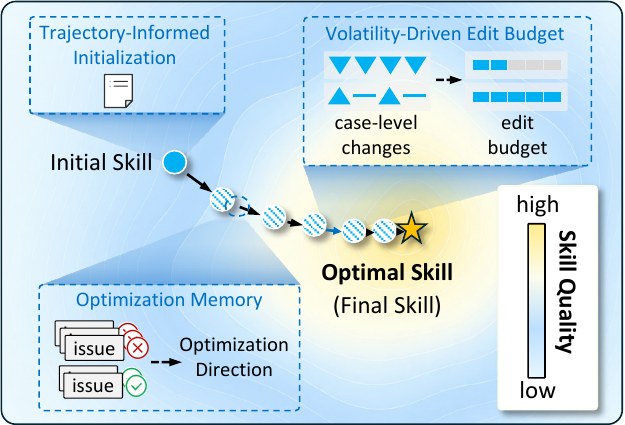}
        \caption{ \sys\ (ours).}
        \label{fig:skilladams_concept}
    \end{subfigure}

    \caption{ 
        Conceptual comparison of iterative skill optimization strategies.
    }
    \label{fig:intro-concept}
\end{figure}

{
To reduce human intervention, recent studies have investigated \emph{automated Skill construction} from interaction experience. AutoManual incrementally updates structured rules and compiles them into an instruction manual~\citep{chen2024automanual}. Agent Skill Induction learns verified programmatic Skills from web interactions~\citep{wang2025asi}, while Trace2Skill synthesizes a unified Skill from a diverse pool of execution traces~\citep{trace2skill}. More recently, several studies have formulated automated Skill construction as \emph{skill self-evolution}. Some methods iteratively refine a Skill document~\mbox{\citep{gautam2026skillaxe,alzubi2026evoskill,skillopt2026}}, while others evolve Skill packages or maintain Skill repositories~\mbox{\citep{zhang2026evoskills,ouyang2026skillos}}. These approaches reduce manual effort by refining Skills over multiple iterations. As illustrated by the iteration-local optimization path in Figure~\ref{fig:intro-concept}(a), however, existing methods still lack reliable control over what to revise and how much to revise in each iteration.
}

{
To address this limitation, we propose \textbf{\sys}, a stable and efficient framework that aims to preserve effective corrections across iterations while adapting each revision to the reliability of recent evidence. Realizing these properties raises two challenges. The first is \textbf{Direction Stability}. Since each iteration observes feedback from only a limited set of cases, successive revisions may focus on different problems and undo one another. For example, one revision may instruct the agent to minimize the total cost by selecting the cheapest feasible itinerary. A later revision may add more attractions to produce a richer travel plan, but these additions can increase the cost and violate the earlier budget constraint. New revisions must therefore remain consistent with effective corrections accumulated in earlier iterations. The second challenge is \textbf{Update Adaptivity}. The appropriate edit scope should depend on how consistently a recent revision affects the evaluated cases. If one revision improves some cases but degrades others, a broad subsequent edit risks overwriting useful guidance and should therefore be constrained. If the gains are consistent across cases, a broader edit is better supported.
}

{
These challenges resemble those in stochastic optimization, where each update is based on partial evidence and the appropriate step size depends on the scale of recent signals. Adam \citep{kingma2015adam} provides two complementary design principles: aggregate historical update signals to stabilize direction, and rescale the effective step size using accumulated signal magnitude. \sys adopts these principles functionally in the discrete skill space rather than applying Adam numerically.
}

{
As illustrated in Figure~\ref{fig:intro-concept}(b), \sys first constructs a trajectory-informed initial skill from execution trajectories and their evaluation feedback. During self-evolution, an \emph{Evolving Issue Tracker} records identified problems, their current status, and the outcomes of prior solution attempts, allowing each new update to consider both current feedback and accumulated evidence. In parallel, a \emph{volatility-driven edit budget} measures how unevenly a candidate changes case-level performance and controls the allowable scope of the next modification. The tracker determines what should be revised, while the budget determines how much may be revised. Together with trajectory-informed initialization, these components produce the more coherent and efficient optimization path shown in Figure~\ref{fig:intro-concept}(b).
}

Our contributions are summarized as follows:
\begin{itemize}[leftmargin=*, topsep=2pt, itemsep=2pt, partopsep=0pt, parsep=0pt]
    \item We propose \textbf{\sys}, an Adam-inspired framework for stable and efficient skill self-evolution.
    
    \item We introduce an \emph{optimization memory} and a \emph{volatility-driven edit budget} as functional analogues of Adam's first- and second-moment mechanisms, respectively stabilizing the optimization direction and adapting the update magnitude.
    
    \item We conduct extensive experiments on seven benchmarks, where \sys achieves state-of-the-art performance while requiring substantially fewer optimization iterations and reducing overall optimization cost.
\end{itemize}

\begin{figure}[t]
    \centering

    \begin{subfigure}[t]{\linewidth}
        \centering
        \includegraphics[width=\linewidth]{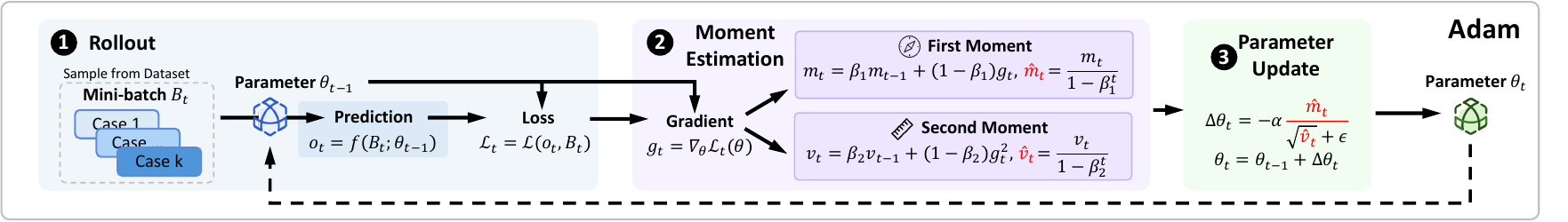}
        \caption{Adam inspiration. Adam aggregates first- and second-moment estimates to
        stabilize the update direction and adapt the effective step size.}
        \label{fig:framework-adam}
    \end{subfigure}

    \vspace{0.5em}

    \begin{subfigure}[t]{\linewidth}
        \centering
        \includegraphics[width=\linewidth]{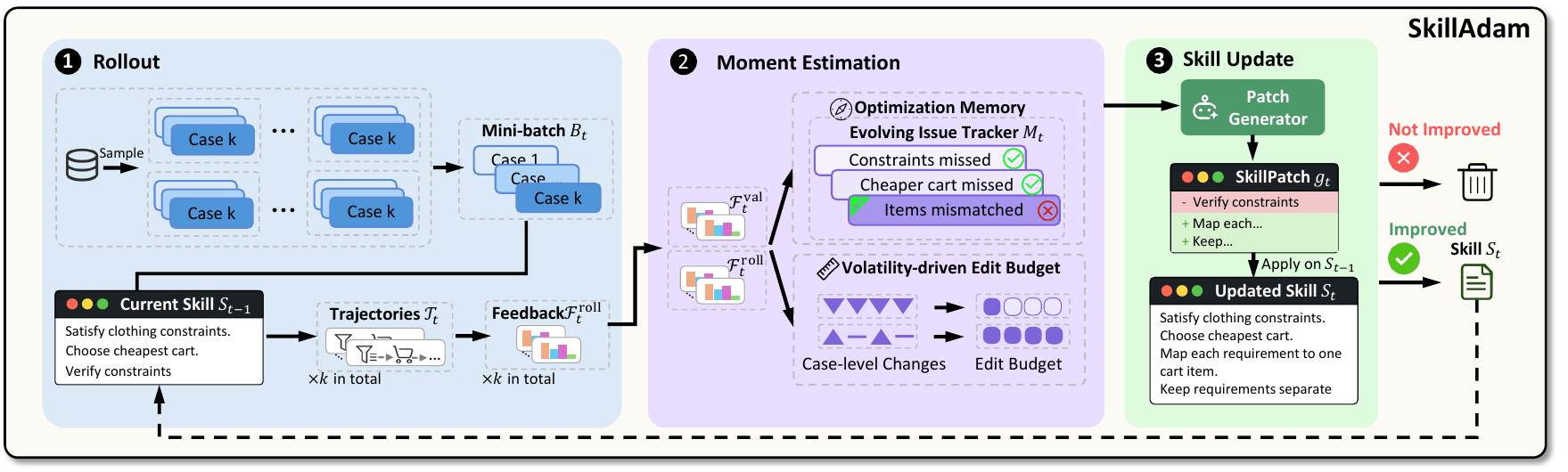}
        \caption{\sys framework. \sys uses the Evolving Issue Tracker and a volatility-driven edit budget to provide analogous control over the direction and magnitude of skill updates in the discrete skill space.}
        \label{fig:framework-sys}
    \end{subfigure}

    \caption{
        Functional correspondence between Adam and \sys.
    }
    \label{fig:framework}
\end{figure}

\section{Related Work}
\label{sec:related-work}

\paragraph{Agent Skills.}
We follow Anthropic's official definition of \emph{Agent Skills} as modular packages of instructions and supporting resources that equip an agent with specialized capabilities~\citep{anthropic2025skills}. Because Skills are stored independently of model parameters, the same package can be distributed to compatible agents without retraining. Earlier work had already externalized reusable capabilities, although it did not share this package definition. Voyager stores executable programs in a library that can be retrieved for later tasks~\citep{wang2023voyager}. AutoManual learns structured rules through interaction and compiles them into a readable instruction manual~\citep{chen2024automanual}. Agent Skill Induction learns and verifies programmatic skills for web agents~\citep{wang2025asi}. SkillsBench provides a common evaluation of package-based Agent Skills and shows that curated Skills improve agent performance across diverse tasks~\citep{skillsbench2026}. ExpeL and Agent Workflow Memory instead retain natural-language experience or reusable workflows from past executions~\citep{zhao2023expel,wang2025awm}. Our work focuses on Skills whose primary artifact is a natural-language instruction document and studies their iterative optimization under task evaluation.

\paragraph{Prompt Optimization.}
Automatic prompt optimization studies how to improve discrete language artifacts without updating model parameters. OPRO proposes new instructions from previously evaluated candidates and their scores~\citep{yang2024opro}. ProTeGi turns error feedback into textual gradients and applies search to select prompt edits~\citep{pryzant2023protegi}. TextGrad propagates natural-language feedback through a computation graph to optimize prompts and other textual components~\citep{textgrad}. GEPA uses reflective feedback from rollouts to evolve prompts~\citep{gepa}. ERM retains feedback from earlier attempts to support exemplar-guided prompt optimization~\citep{yan2025erm}. They show that evaluation signals can guide discrete textual updates without changing model parameters. Their optimization targets are prompts or components of language-model programs, while \sys optimizes a reusable Agent Skill used across task instances.

\paragraph{Skill Self-Evolution.}
Recent work directly automates Agent Skill construction and revision. Trace2Skill analyzes a broad pool of executions and consolidates trajectory-local lessons into a unified skill directory~\citep{trace2skill}. SkillAxe iteratively diagnoses and refines LLM-authored skill documents with structured evaluation signals~\citep{gautam2026skillaxe}. EvoSkill discovers and revises skills through failure analysis, then retains validated candidates through Pareto selection~\citep{alzubi2026evoskill}. CoEvoSkills jointly evolves a Skill Generator and a Surrogate Verifier to construct multi-file Skill packages without ground-truth test content~\citep{zhang2026evoskills}. SkillOS trains a curator that updates an external skill repository from accumulated experience~\citep{ouyang2026skillos}. SkillOpt is closest to our setting because it applies bounded textual edits to one skill document and accepts an update only when validation performance improves~\citep{skillopt2026}. Their optimization targets differ. Some optimize one document, while others construct packages or maintain a repository. Across these settings, existing methods do not jointly maintain persistent optimizer states for the direction and magnitude of successive revisions.

{
\section{Preliminaries}

\label{sec:preliminary}

\subsection{Skill Optimization Problem}

We consider a skill as a Markdown-formatted natural-language instruction that guides an agent's behavior in a target domain. In this work, we focus on Skills whose primary artifact is such a structured natural-language instruction document. Let $\mathcal{S}$ denote the discrete space of possible skills and $\mathcal{D}=\{d_1,\ldots,d_N\}$ denote a task dataset.
}

Given a skill $S\in\mathcal{S}$ and a task $d\in\mathcal{D}$, the domain evaluator returns structured case-level evaluation feedback
\begin{equation}
\mathcal{E}(S,d)
=
\left(
E(S,d),
C(S,d)
\right),
\label{eq:evaluation-feedback}
\end{equation}
{
where $E(S,d)\in\mathbb{R}^{m}$ is an $m$-dimensional vector of domain-specific metrics, and $C(S,d)$ denotes optional diagnostic information, such as error descriptions or judge rationales. The metric vector $E$ supports numerical aggregation and acceptance decisions, whereas $C$ is retained in $\mathcal{F}$ as language-space evidence for patch generation and issue tracking. We use
}
\begin{equation}
\mathcal{F}(S,B)
=
\left\{
\mathcal{E}(S,d)
\mid
d\in B
\right\}
\label{eq:batch-feedback}
\end{equation}
to denote the evaluation feedback collected over a task batch $B\subseteq\mathcal{D}$.

Let $\Phi$ be a task-dependent aggregation function that maps task-level metric vectors to a scalar objective. We define
\begin{equation}
J(S)
=
\Phi
\left(
\left\{
E(S,d)
\mid
d\in\mathcal{D}
\right\}
\right),
\qquad
S^{*}
=
\arg\max_{S\in\mathcal{S}}J(S).
\label{eq:skill-objective}
\end{equation}
{
The target domain determines the precise forms of $E$ and $C$, and its evaluation protocol determines $\Phi$. Equation~\ref{eq:skill-objective} aggregates only $E$ because $J$ is numerical; the diagnostic content $C$ remains available through $\mathcal{F}$ to the LLM-based update and memory functions.
}

{
\subsection{Adam Optimization}
}
In differentiable optimization, Adam \citep{kingma2015adam} maintains exponential moving averages of the first and second moments of stochastic gradients. Let
$\nabla_t=\nabla_{\theta}\mathcal{L}_t(\theta_{t-1})$
denote the stochastic gradient of the mini-batch loss at iteration $t$. Adam updates
\begin{align}
m_t^{\mathrm{Adam}}
&=
\beta_1m_{t-1}^{\mathrm{Adam}}
+
(1-\beta_1)\nabla_t,
\label{eq:adam-first-moment}
\\
v_t^{\mathrm{Adam}}
&=
\beta_2v_{t-1}^{\mathrm{Adam}}
+
(1-\beta_2)\nabla_t^2,
\label{eq:adam-second-moment}
\end{align}
Here, $m_t^{\mathrm{Adam}}$ and $v_t^{\mathrm{Adam}}$ are the first- and second-moment estimates. The coefficients $\beta_1,\beta_2\in[0,1)$ are exponential decay rates. The square is applied element-wise. Since both moment estimates are initialized at zero, Adam applies bias correction:
\begin{equation}
\widehat{m}_t^{\mathrm{Adam}}
=
\frac{m_t^{\mathrm{Adam}}}{1-\beta_1^t},
\qquad
\widehat{v}_t^{\mathrm{Adam}}
=
\frac{v_t^{\mathrm{Adam}}}{1-\beta_2^t}.
\label{eq:adam-bias-correction}
\end{equation}
The parameters are then updated as
\begin{equation}
\theta_t
=
\theta_{t-1}
-
\alpha
\frac{\widehat{m}_t^{\mathrm{Adam}}}
{\sqrt{\widehat{v}_t^{\mathrm{Adam}}}+\epsilon},
\label{eq:adam-update}
\end{equation}
where $\alpha>0$ is the base learning rate and $\epsilon>0$ is a small constant that prevents division by zero and improves numerical stability. The first-moment estimate aggregates gradient information across iterations to stabilize the update direction, while the second-moment estimate rescales the base learning rate according to the recent squared-gradient magnitude, yielding an adaptive effective step size. The ratio
\begin{equation}
\alpha_t^{\mathrm{eff}}
=
\frac{\alpha}
{\sqrt{\widehat{v}_t^{\mathrm{Adam}}}+\epsilon}
\label{eq:adam-effective-step}
\end{equation}
can be interpreted as the element-wise effective step size.

\section{Method}
\label{sec:method}

To enable stable and efficient skill self-evolution in a discrete skill space,
we propose \textbf{\sys} with two Adam-inspired mechanisms: an Evolving
Issue Tracker that stabilizes the update direction and a volatility-driven
edit budget that adapts the update magnitude. We describe the framework and its
components below.

\subsection{The \sys Framework}
\label{sec:sys-framework}

\paragraph{Framework Overview.}
Figure~\ref{fig:framework} presents the \sys cycle. It begins with \emph{rollout}, after which \emph{moment estimation} informs the \emph{skill update}. Before iterative optimization begins, \sys constructs an initial skill $S_0$ from execution trajectories and their evaluation feedback. At iteration $t$, the current skill is executed on a sampled mini-batch, and the optimizer states carried from the previous iteration, $M_{t-1}$ and $\sigma_t$, guide the generation of a candidate modification. After the candidate has been evaluated, the \emph{Evolving Issue Tracker} compares the rollout and validation outcomes and evolves the structured issue list that constitutes the optimization memory $M_t$. The same paired feedback is used to update the improvement-volatility estimate.

The three blocks in Figure~\ref{fig:framework} describe functional roles. They do not prescribe a strict within-iteration execution order. In particular, the memory and volatility states obtained from the validation outcome of iteration $t$ guide the skill update at iteration $t+1$. The complete operational order is given in Algorithm~\ref{alg:sys}, while Table~\ref{tab:adam-sys-correspondence} summarizes the functional correspondence between Adam and \sys.

\paragraph{Rollout.}
At iteration $t$, \sys samples a mini-batch $B_t\subset\mathcal{D}$ and executes the agent with the current skill $S_{t-1}$:
\begin{equation}
\mathcal{T}_t
=
\operatorname{Rollout}
\left(
S_{t-1},
B_t
\right),
\label{eq:trajectory-collection}
\end{equation}
where $\mathcal{T}_t$ denotes the collected execution trajectories. The domain evaluator then produces the corresponding case-level evaluation feedback
\begin{equation}
\mathcal{F}_t^{\mathrm{roll}}
=
\mathcal{F}
\left(
S_{t-1},
B_t
\right),
\label{eq:rollout-feedback}
\end{equation}
as defined in Equation~\ref{eq:batch-feedback}. The trajectories record how the agent behaves during execution, while $\mathcal{F}_t^{\mathrm{roll}}$ contains the resulting metric scores and diagnostic information. Together, $(\mathcal{T}_t,\mathcal{F}_t^{\mathrm{roll}})$ provide the iteration-specific optimization signal used to generate the next skill update.

\paragraph{Moment Estimation.}
\sys maintains two persistent optimizer states across iterations. The
optimization memory is instantiated as an \emph{Evolving Issue Tracker}
(EIT), denoted by $M_t$, while the volatility estimate $V_t$ controls the
magnitude of subsequent updates. Both states are refreshed after the candidate
skill has been evaluated and are then carried into the next iteration.

The Evolving Issue Tracker is represented as a collection of structured issue
entries:
\begin{equation}
M_t
=
\left\{
I_j
\mid
j\in\mathcal{J}_t
\right\},
\qquad
I_j
=
\left(
p_j,
z_j,
\mathcal{A}_j
\right),
\label{eq:memory-structure}
\end{equation}
where $\mathcal{J}_t$ is the set of issue identifiers recorded through
iteration $t$. The pair $(p_j,z_j)$ describes an error pattern and its current
status. The set $\mathcal{A}_j$ stores previous solution attempts with their
observed outcomes. The tracker therefore maintains each issue and its
resolution history.

After the candidate skill at iteration $t$ has been evaluated, the tracker
directly evolves its state:
\begin{equation}
M_t
=
\mathcal{U}_{\mathrm{EIT}}
\left(
M_{t-1},
\mathcal{T}_t,
\mathcal{F}_t^{\mathrm{roll}},
g_t,
\mathcal{F}_t^{\mathrm{val}},
a_t
\right).
\label{eq:memory-update}
\end{equation}
The LLM-based update function $\mathcal{U}_{\mathrm{EIT}}$ compares the
rollout and validation outcomes to maintain the issue records. It links
observed failures to existing issues when possible and creates entries when
needed. It also records each attempted modification with its outcome and
reopens an issue when the same failure recurs.

The Evolving Issue Tracker serves as a functional analogue of Adam's
first-moment estimate:
\begin{equation}
m_t^{\mathrm{Adam}}
\quad\longleftrightarrow\quad
M_t.
\label{eq:first-moment-correspondence}
\end{equation}
Both states integrate current evidence with information accumulated across
previous iterations. Adam aggregates numerical gradients. The Evolving Issue
Tracker instead accumulates issue histories and solution outcomes to stabilize
the update direction.

\sys additionally estimates how consistently the candidate affects the
evaluated cases. Let
$\mathcal{E}_{t,i}^{\mathrm{roll}}$ and
$\mathcal{E}_{t,i}^{\mathrm{val}}$ denote the case-level feedback for the
current and candidate skills, respectively, on case $d_i\in B_t$. The
case-level improvement is
\begin{equation}
\delta_{t,i}
=
s\!\left(
\mathcal{E}_{t,i}^{\mathrm{val}}
\right)
-
s\!\left(
\mathcal{E}_{t,i}^{\mathrm{roll}}
\right),
\qquad
d_i\in B_t,
\label{eq:improvement-delta}
\end{equation}
where $s(\cdot)$ extracts the benchmark-specific scalar score used to measure
improvement. The mean case-level improvement is
\begin{equation}
\overline{\delta}_t
=
\frac{1}{|B_t|}
\sum_{d_i\in B_t}
\delta_{t,i}.
\label{eq:mean-improvement}
\end{equation}
\sys estimates the current improvement volatility using
\begin{equation}
\widehat{V}_t
=
\frac{1}{|B_t|-1}
\sum_{d_i\in B_t}
\left(
\delta_{t,i}
- 
\overline{\delta}_t
\right)^2,
\qquad
|B_t|\geq 2.
\label{eq:instantaneous-volatility}
\end{equation}
When fewer than two valid case-level comparisons are available, we set
$\widehat{V}_t=0$. A high value indicates that the candidate produces
substantially different effects across cases, while a low value indicates
more consistent effects.

The history-weighted volatility estimate is updated as
\begin{equation}
V_t
=
\beta_2V_{t-1}
+
(1-\beta_2)\widehat{V}_t,
\qquad
V_0=0.
\label{eq:volatility-ema}
\end{equation}
The edit budget for the next iteration is then computed by
\begin{equation}
\sigma_{t+1}
=
\max
\left(
b_{\min},
\left\lfloor
b_{\mathrm{base}}
\left[
1-
\operatorname{clip}
\left(
\frac{V_t}{V_{\max}},
0,
1
\right)
\right]
\right\rceil
\right),
\label{eq:edit-budget}
\end{equation}
where $b_{\mathrm{base}}$ is the base edit budget and $b_{\min}$ is the
minimum allowable budget. The volatility saturation threshold is $V_{\max}$. High volatility yields a smaller edit budget, while low volatility
permits a broader modification. The budget used at iteration $t$,
$\sigma_t$, is determined from the state accumulated through iteration
$t-1$.

This mechanism provides the following correspondence:
\begin{equation}
v_t^{\mathrm{Adam}}
\quad\longleftrightarrow\quad
V_t,
\qquad
\alpha_t^{\mathrm{eff}}
\quad\longleftrightarrow\quad
\sigma_{t+1}.
\label{eq:second-moment-correspondence}
\end{equation}
The optimization memory therefore determines which problems should guide the
subsequent update, while the volatility-driven edit budget determines how
extensively the skill may be modified.
\paragraph{Skill Update.}
At iteration $t$, the LLM-based patch generator proposes a skill modification
conditioned on the current rollout evidence and the optimizer states carried
from the previous iteration:
\begin{equation}
g_t
=
\mathcal{G}_{\mathrm{LLM}}
\left(
S_{t-1},
\mathcal{T}_t,
\mathcal{F}_t^{\mathrm{roll}},
M_{t-1},
\sigma_t
\right).
\label{eq:patch-generation}
\end{equation}
Here, $\mathcal{T}_t$ and $\mathcal{F}_t^{\mathrm{roll}}$ describe the
behavior observed in the current iteration, $M_{t-1}$ provides
cross-iteration guidance on which problems should be addressed, and
$\sigma_t$ controls the allowable scope of the modification.

Applying the generated modification to the current skill produces a candidate:
\begin{equation}
\widetilde{S}_t
=
\operatorname{Apply}
\left(
S_{t-1},
g_t
\right).
\label{eq:candidate-skill}
\end{equation}
The candidate is evaluated on the same mini-batch used for the current-skill
rollout:
\begin{equation}
\mathcal{F}_t^{\mathrm{val}}
=
\mathcal{F}
\left(
\widetilde{S}_t,
B_t
\right).
\label{eq:validation-feedback}
\end{equation}
Consequently, $\mathcal{F}_t^{\mathrm{roll}}$ and
$\mathcal{F}_t^{\mathrm{val}}$ provide directly comparable case-level
feedback for the current and candidate skills.

A benchmark-specific acceptance gate determines whether the candidate should
replace the current skill:
\begin{equation}
a_t
=
G_{\theta}
\left(
S_{t-1},
\widetilde{S}_t,
\mathcal{F}_t^{\mathrm{roll}},
\mathcal{F}_t^{\mathrm{val}}
\right)
\in
\{0,1\}.
\label{eq:acceptance-gate}
\end{equation}
The skill is updated as
\begin{equation}
S_t
=
\begin{cases}
\widetilde{S}_t, & a_t=1,\\
S_{t-1}, & a_t=0.
\end{cases}
\label{eq:skill-update}
\end{equation}

The Evolving Issue Tracker uses each attempted modification and its evaluated
outcome to evolve from $M_{t-1}$ to $M_t$. The same paired feedback is used
to compute $\widehat{V}_t$ and $V_t$. The updated skill and optimizer state
are carried into the next iteration, closing the optimization loop shown in
Figure~\ref{fig:framework}.
\subsection{Algorithm and Adam Correspondence}
\label{sec:algorithm}

Algorithm~\ref{alg:sys} presents the operational order of \sys.
The current skill first produces rollout trajectories and evaluation feedback
on a sampled mini-batch. The optimizer states carried from the previous
iteration then guide the generation of a candidate skill. After the candidate
has been evaluated on the same cases, \sys decides whether to accept it
and uses the resulting comparison to evolve the Evolving Issue Tracker and
update the volatility estimate. These updated states guide the next
iteration.

\begin{algorithm}[t]
\caption{\sys: Adam-Inspired Skill Self-Evolution}
\label{alg:sys}
\begin{algorithmic}[1]

{
\REQUIRE Task dataset $\mathcal{D}$, evaluator $\mathcal{E}$, mini-batch size $k$,
maximum iterations $T_{\max}$, EMA coefficient $\beta_2$, budget parameters
$b_{\mathrm{base}}, b_{\min}, V_{\max}$, and random seed $\xi$
}

\ENSURE Optimized skill $S_{T_{\max}}$

{
\STATE $(\mathcal{T}_0,\mathcal{F}_0)
\leftarrow
\operatorname{CollectInitializationData}
(\mathcal{D},\mathcal{E},\xi)$
}

\STATE $S_0
\leftarrow
\operatorname{Initialize}_{\mathrm{LLM}}
(\mathcal{T}_0,\mathcal{F}_0)$

\STATE $M_0\leftarrow\emptyset$

\STATE $V_0\leftarrow0$

\STATE $\sigma_1\leftarrow b_{\mathrm{base}}$

\FOR{$t=1,2,\ldots,T_{\max}$}

    \STATE $B_t \leftarrow \operatorname{Sample}(\mathcal{D},k,\xi+t),\quad
    \mathcal{T}_t \leftarrow \operatorname{Rollout}(S_{t-1},B_t),\quad
    \mathcal{F}_t^{\mathrm{roll}} \leftarrow \mathcal{F}(S_{t-1},B_t)$

    \STATE $g_t \leftarrow
    \mathcal{G}_{\mathrm{LLM}}
    (S_{t-1},\mathcal{T}_t,\mathcal{F}_t^{\mathrm{roll}},M_{t-1},\sigma_t),\quad
    \widetilde{S}_t \leftarrow \operatorname{Apply}(S_{t-1},g_t)$

    \IF{$\widetilde{S}_t=\mathbf{null}$}

        \STATE $S_t\leftarrow S_{t-1},\quad
        M_t\leftarrow M_{t-1},\quad
        V_t\leftarrow V_{t-1},\quad
        \sigma_{t+1}\leftarrow\sigma_t$

        \STATE \textbf{continue}

    \ENDIF

    \STATE $\mathcal{F}_t^{\mathrm{val}} \leftarrow
    \mathcal{F}(\widetilde{S}_t,B_t),\quad
    a_t \leftarrow
    G_{\theta}(S_{t-1},\widetilde{S}_t,
    \mathcal{F}_t^{\mathrm{roll}},\mathcal{F}_t^{\mathrm{val}})$

    \IF{$a_t=1$}

        \STATE $S_t\leftarrow\widetilde{S}_t$

    \ELSE

        \STATE $S_t\leftarrow S_{t-1}$

    \ENDIF

    \STATE $M_t
    \leftarrow
    \mathcal{U}_{\mathrm{EIT}}
    \left(
    M_{t-1},
    \mathcal{T}_t,
    \mathcal{F}_t^{\mathrm{roll}},
    g_t,
    \mathcal{F}_t^{\mathrm{val}},
    a_t
    \right)$

    \STATE $\{\delta_{t,i}\}_{d_i\in B_t}
    \leftarrow
    \Delta
    \left(
    \mathcal{F}_t^{\mathrm{roll}},
    \mathcal{F}_t^{\mathrm{val}}
    \right)$

    \STATE $\widehat{V}_t
    \leftarrow
    \operatorname{Var}
    \left(
    \{\delta_{t,i}\}_{d_i\in B_t}
    \right)$

    \STATE $V_t
    \leftarrow
    \beta_2V_{t-1}
    +
    (1-\beta_2)\widehat{V}_t$

    \STATE $\sigma_{t+1}
    \leftarrow
    \max
    \left(
    b_{\min},
    \left\lfloor
    b_{\mathrm{base}}
    \left[
    1-
    \operatorname{clip}
    \left(
    \frac{V_t}{V_{\max}},
    0,
    1
    \right)
    \right]
    \right\rceil
    \right)$

\ENDFOR

\RETURN $S_{T_{\max}}$

\end{algorithmic}
\end{algorithm}

Figure~\ref{fig:framework} provides a block-level comparison between Adam and
\sys, while Table~\ref{tab:adam-sys-correspondence} makes the
correspondence between their optimization quantities explicit. The analogy is functional, not numerical. \sys does not compute gradients
in the discrete skill space. It constructs language-space states that serve
the same optimization roles.

\begin{table}[t]
\centering
\caption{
Core functional correspondence between Adam and \sys.
The correspondence is functional, with evaluation feedback and a language-space
patch serving the roles of the loss and gradient.
}
\label{tab:adam-sys-correspondence}
\small
\setlength{\tabcolsep}{4.5pt}
\begin{tabular}{lll}
\toprule
Adam
& \sys
& Functional role \\
\midrule

Parameters $\theta_{t-1}$
& Current skill $S_{t-1}$
& Represent the optimized state \\

Prediction $o_t$
& Execution trajectories $\mathcal{T}_t$
& Record the execution output \\

Loss $\mathcal{L}_t$
& Evaluation feedback $\mathcal{F}^{\mathrm{roll}}_t$
& Evaluate the current behavior \\

Gradient $\nabla_t$
& Skill patch $g_t$
& Provide the local update signal \\

First moment $m_t^{\mathrm{Adam}}$
& Evolving Issue Tracker $M_t$
& Stabilize the update direction \\

Second moment $v_t^{\mathrm{Adam}}$
& Historical volatility $V_t$
& Accumulate update variability \\

Effective step size $\alpha_t^{\mathrm{eff}}$
& Edit budget $\sigma_{t+1}$
& Adapt the update magnitude \\

Apply $\Delta\theta_t$
& Apply and gate $g_t$
& Update the optimized state \\

Updated parameter $\theta_t$
& Accepted skill $S_t$
& Carry the state forward \\

\bottomrule
\end{tabular}
\end{table}

The rollout feedback $\mathcal{F}_t^{\mathrm{roll}}$ plays a loss-like role by
evaluating the current behavior, while the generated patch $g_t$ serves as the
gradient-like local update signal in the skill space. The Evolving Issue Tracker
$M_t$ accumulates issue histories and solution outcomes across iterations,
analogous to Adam's first moment. Similarly, $V_t$ aggregates recent improvement
variability and determines the next edit budget $\sigma_{t+1}$.
\section{Experiments}
\label{sec:exp}

\subsection{Benchmarks}
\label{sec:exp-benchmarks}

We evaluate \sys on seven benchmarks that cover knowledge work and interactive planning. Six benchmarks contribute one evaluation slice each. DeepPlanning contributes Shopping Levels 1--3 and Travel EN, giving ten slices in total. Each benchmark uses its native task environment, tool interface, and evaluator.

We classify five benchmarks as \emph{short-horizon}, as they require fewer than ten tool calls per task on average. SearchQA~\citep{searchqa} evaluates open-domain question answering with noisy retrieved context. SpreadsheetBench~\citep{spreadsheetbench} requires an agent to modify spreadsheets from natural-language instructions. OfficeQA~\citep{officeqa} evaluates question answering over historical U.S. Treasury Bulletins. DocVQA~\citep{docvqa} evaluates visual question answering over document images. LiveMathematicianBench (LMB; also abbreviated as LiveMath in the tables)~\citep{livemathbench} evaluates reasoning over mathematical theorems and proof sketches. The two \emph{long-horizon} benchmarks require at least ten tool calls per task on average. ALFWorld~\citep{alfworld} contains text-based embodied household tasks. DeepPlanning~\citep{zhang2026deepplanning} evaluates multi-step shopping and travel planning.

\paragraph{Metrics.}
We report the standard primary metric for each benchmark. SearchQA, OfficeQA, and LMB use Exact Match. SpreadsheetBench uses hard task success, which requires the complete spreadsheet-editing task to pass. DocVQA uses ANLS-hard. ALFWorld uses episode goal-completion rate. DeepPlanning reports Shopping Case Accuracy and Travel Case Accuracy. DP-Shopping pools the three shopping levels by case count, with $25$, $25$, and $10$ test cases. DP-Travel contains $60$ test cases. We define $\mathrm{DP\text{-}Avg}=(\mathrm{DP\text{-}Shopping}+\mathrm{DP\text{-}Travel})/2$ using the unrounded domain-level accuracies. It is not an equal average of the four DeepPlanning slices. All main results are percentages.

\subsection{Baselines}
\label{sec:exp-baselines}

We compare \sys with seven baselines. \textbf{NoSkill} runs the target agent without an injected skill. \textbf{HumanSkill} uses an expert-authored skill, and \textbf{LLMSkill} uses a skill written in one LLM call. The optimization baselines are \textbf{Trace2Skill}~\citep{trace2skill}, \textbf{TextGrad}~\citep{textgrad}, \textbf{GEPA}~\citep{gepa}, and \textbf{SkillOpt}~\citep{skillopt2026}.

For the five short-horizon benchmarks, baseline results are taken from the GPT-5.5 no-harness setting reported by SkillOpt. For ALFWorld, the NoSkill result is taken from SkillOpt, and we reproduce the SkillOpt result in our environment. For DeepPlanning, we reproduce both NoSkill and SkillOpt in the task environment used for \sys. Within each benchmark, the compared methods use the same target-agent configuration, test cases, and evaluator. \sys and SkillOpt also start from the same initial skill, which is constructed from a fixed set of baseline execution trajectories.

SkillOpt retains its original train and selection split together with its native selection, slow-update, and optimizer-memory procedures. We run SkillOpt for four epochs under this protocol.

\subsection{Setup}
\label{sec:exp-setup}

\paragraph{Common Protocol.}
All experiments use a no-harness, direct-chat setting. The agent interacts with the native task interface and evaluator of each benchmark without an additional orchestration layer. Skills are injected as natural-language instructions, and the target model remains frozen during optimization. We use GPT-5.5 for the six benchmarks outside DeepPlanning. These runs use medium reasoning effort, a temperature of $1.0$, and a maximum output length of $16{,}384$ tokens. DeepPlanning uses Claude Sonnet 4.5 with a temperature of $0.0$ and the same output limit. Travel EN uses a separate frozen model to convert the generated plan into the required format before evaluation.

\paragraph{Initialization and Optimization.}
\sys and SkillOpt share an initial skill constructed from fixed baseline execution trajectories. For the six benchmarks outside DeepPlanning, \sys merges the original train and selection partitions into one optimization pool. SkillOpt retains the original split. The test partition is unchanged and is used only for final evaluation. DeepPlanning uses an odd-even split by case identifier. Odd-numbered cases are used for optimization, and even-numbered cases are reserved for testing. We use seed $42$ for local case sampling, optimization-pool shuffling, and other controlled random operations.

On the six benchmarks outside DeepPlanning, \sys makes one pass through the optimization pool and uses a benchmark-specific stopping rule. DeepPlanning samples optimization batches at random and uses task-specific stopping criteria. At each iteration, \sys evaluates the current skill and its proposed revision on the same sampled cases. Their execution trajectories and evaluation feedback are passed to the optimizer states described in Section~\ref{sec:sys-framework}.

\paragraph{Acceptance Protocol.}
The acceptance gate uses the benchmark's primary and auxiliary metrics. A proposed revision is accepted when at least one designated metric reaches its improvement threshold and every protected metric remains within its regression boundary. An auxiliary metric can therefore support acceptance when the primary metric is unchanged. \sys does not use an additional validation set for this decision. SkillOpt follows its native selection and slow-update rules.

\paragraph{Evaluation.}
Each test case is evaluated once with the target-agent configuration specified above.

\subsection{Main Results}
\label{sec:exp-main}

We compare \sys with all baselines on two categories of benchmarks, i.e., \emph{short-horizon} benchmarks and \emph{long-horizon} benchmarks. The results are recorded in Table~\ref{tab:single} and Table~\ref{tab:agentic}.

{
\paragraph{Results on Short-Horizon Benchmarks.}
As illustrated in Table~\ref{tab:single},
\sys achieves the overall best performance among all baselines, with four strict wins and one tie on five benchmarks. Specifically, \sys achieves remarkably better performance compared with HumanSkill and LLMSkill, with an average improvement of 14.45\% and 31.20\% respectively. This is because \sys adopts iterative skill optimization instead of one-shot skill generation. Thus, it can optimize the skill based on the evaluation result of the immediate agent rollouts, which produces higher-quality skills.
In addition, compared with the second-best methods, i.e., SkillOpt, \sys still shows better performance. For example, \sys achieves higher accuracy on DocVQA and LiveMath, with an improvement of 1.21\% and 1.20\%. This improvement stems from two technical designs of \sys, i.e., Optimization Memory and Volatility-driven Edit Budget. With these designs, we can optimize the skills more stably. Thus, we can produce better skills than SkillOpt.
}

\begin{table}[t]
\centering
\caption{Main results on five short-horizon benchmarks. Baselines follow the GPT-5.5
no-harness results reported by SkillOpt. Scores are percentages. Bold and
underlining mark the best and second-best values.}
\label{tab:single}
\small
\begin{tabular}{lccccc}
\toprule
Method & SearchQA & Spreadsheet & OfficeQA & DocVQA & LiveMath \\
\midrule
NoSkill     & 77.7 & 41.8 & 33.1 & 78.8 & 37.6 \\
HumanSkill  & 81.8 & 72.9 & 66.9 & 90.1 & 38.4 \\
LLMSkill    & 80.9 & 43.2 & 51.7 & 89.6 & 40.0 \\
Trace2Skill  & 82.4 & 49.6 & 65.7 & 90.6 & 52.0 \\
TextGrad     & 81.4 & 41.1 & 42.0 & 87.2 & 49.2 \\
GEPA         & 84.8 & 73.6 & 63.9 & 89.1 & 43.2 \\
SkillOpt     & \underline{87.3} & \underline{80.7} & \textbf{72.1} & \underline{91.2} & \underline{66.9} \\
\midrule
\textbf{\sys} & \textbf{87.5} & \textbf{81.1} & \textbf{72.1} & \textbf{92.3} & \textbf{67.7} \\
\bottomrule
\end{tabular}
\end{table}

\paragraph{Results on Long-Horizon Benchmarks.}
{
Table~\ref{tab:agentic} reports the long-horizon results. As illustrated in the table, the advantages of \sys are more prominently demonstrated. Specifically, while both NoSkill and SkillOpt nearly fail on DP-Travel, \sys achieves an accuracy of 11.7\%. Moreover, compared with SkillOpt, \sys improves DP-Avg from 21.7\% to 28.3\%, a gain of 6.7 percentage points computed from the unrounded accuracies. The results demonstrated that our proposed evolution algorithm can work better on more challenging tasks.
}

In summary, \sys achieves the best overall performance compared with other baselines, demonstrating the effectiveness of the proposed skill evolution strategy. 

\begin{table}[t]
\centering
\caption{Main results on ALFWorld with GPT-5.5 and DeepPlanning with Claude Sonnet 4.5.
DP-Avg averages the unrounded DP-Shopping and DP-Travel accuracies. Scores are
percentages. Bold and underlining mark the best and second-best values.}
\label{tab:agentic}
\small
\begin{tabular}{lcccc}
\toprule
Method & ALFWorld & DP-Shopping & DP-Travel & DP-Avg \\
\midrule
NoSkill & 83.6 & 31.7 & 0.0 & 15.8 \\
SkillOpt & \underline{87.3} & \underline{41.7} & \underline{1.7} & \underline{21.7} \\
\midrule
\textbf{\sys} & \textbf{89.6} & \textbf{45.0} & \textbf{11.7} & \textbf{28.3} \\
\bottomrule
\end{tabular}
\end{table}

\subsection{Ablation Studies}
\label{sec:exp-ablation}

{
We conduct a cumulative ablation study on DeepPlanning to examine the two core mechanisms. Starting from the full framework, we first remove the volatility-driven edit budget and then additionally remove the optimization memory. Thus, each row removes one additional component from the configuration above it. All variants follow the same training and evaluation protocol.
}

\begin{table}[t]
\centering
\caption{Cumulative ablation on DeepPlanning. $\mathsf{M}$ and $\mathsf{B}$ denote the
optimization memory and volatility-driven edit budget. Each row removes one
additional component. Scores are percentages, and bold marks the best value
in each column.}
\label{tab:ablation}
\small
\setlength{\tabcolsep}{5pt}
\begin{tabular}{@{}l@{\hspace{0.8em}}cccc@{\hspace{1em}}c@{\hspace{1em}}c@{}}
\toprule
Configuration & \multicolumn{4}{c}{DP-Shopping} & DP-Travel & DP-Avg \\
\cmidrule(lr){2-5}
              & L1 & L2 & L3 & All & & \\
\midrule
\textbf{\sys ($\mathsf{M}+\mathsf{B}$)}                                     & 52.0 & \textbf{36.0} & \textbf{50.0} & \textbf{45.0} & \textbf{11.7} & \textbf{28.3} \\
\quad $-\mathsf{B}$                       & 48.0 & \textbf{36.0} & 40.0 & 41.7 & 1.7 & 21.7 \\
\quad $-\mathsf{B},-\mathsf{M}$  & \textbf{56.0} & 20.0 & 30.0 & 36.7 & 1.7 & 19.2 \\
NoSkill                                                      & 40.0 & 24.0 & 30.0 & 31.7 & 0.0 & 15.8 \\
\bottomrule
\end{tabular}
\end{table}

{
As illustrated in Table~\ref{tab:ablation}, the full \sys achieves the best overall result, increasing DP-Avg from 19.2\% without both mechanisms to 28.3\%. Specifically, with Optimization Memory fixed, adding the Volatility-driven Edit Budget improves DP-Avg by 6.7 percentage points, with the largest gain occurring on DP-Travel. This result supports the importance of Update Adaptivity in long-horizon planning. Specifically, when a revision produces inconsistent effects across cases, reducing the next edit scope can avoid damaging previously correct constraints, whereas consistent improvements permit broader updates.

Optimization Memory provides a complementary benefit. Without the edit budget, adding memory raises DP-Avg from 19.2\% to 21.7\%, with gains on Shopping L2 and L3 despite a decrease on L1. The concentration of gains on the more difficult levels suggests that recording issue states and prior solution outcomes helps preserve and reconcile multiple corrections across iterations, thereby maintaining a more stable optimization direction. However, because this is a cumulative ablation, it does not independently isolate the effect of memory when the edit budget is enabled or the interaction between the two mechanisms.
}

\subsection{Cross-Model Transfer}
\label{sec:exp-transfer}

We evaluate whether the optimized skills remain effective when transferred to a different agent backbone. Specifically, the skills produced by \sys and SkillOpt using GPT-5.5 are deployed verbatim on GPT-5.4-mini without further optimization or adaptation. We conduct this evaluation on six benchmarks supported by both backbones.

Table~\ref{tab:transfer} reports the task score obtained on GPT-5.4-mini and the corresponding retention ratio. Let $s_{\mathrm{src}}$ and $s_{\mathrm{tgt}}$ denote the scores obtained on GPT-5.5 and GPT-5.4-mini, respectively. We define the retention ratio as $\operatorname{Retention}=s_{\mathrm{tgt}}/s_{\mathrm{src}}$. A higher retention ratio indicates that a larger fraction of the skill's source-model performance is preserved after transfer. No-Skill results on GPT-5.4-mini are included as reference performance and are taken from \citet{skillopt2026}.

\begin{table}[t]
\centering
\caption{Cross-model transfer from GPT-5.5 to GPT-5.4-mini. Score denotes target-model
task performance. Retention measures the percentage of source-model
performance preserved after transfer. Both metrics are percentages. Bold
marks the better result between SkillOpt and \sys.}
\label{tab:transfer}
\small
\setlength{\tabcolsep}{3.2pt}
\begin{tabular}{
    ll
    c
    c
    c
    c
    c
    c
    c
}
\toprule
Method
& Metric
& SearchQA
& DocVQA
& Spreadsheet
& ALFWorld
& LiveMath
& OfficeQA
& Average \\
\midrule

NoSkill
& Score
& 75.9
& 71.4
& 36.1
& 73.1
& 14.7
& 22.1
& 48.9 \\

\midrule

SkillOpt
& Score
& \textbf{80.8}
& 90.4
& 61.4
& 66.4
& 28.2
& 51.2
& 63.1 \\

SkillOpt
& Retention
& \textbf{92.6\%}
& \textbf{99.1\%}
& 76.1\%
& 76.1\%
& 42.2\%
& 71.0\%
& 76.2\% \\

\midrule

\textbf{\sys}
& Score
& 78.9
& \textbf{91.2}
& \textbf{65.0}
& \textbf{82.8}
& \textbf{33.1}
& \textbf{55.8}
& \textbf{67.8} \\

\textbf{\sys}
& Retention
& 90.2\%
& 98.8\%
& \textbf{80.1\%}
& \textbf{92.4\%}
& \textbf{48.9\%}
& \textbf{77.4\%}
& \textbf{81.3\%} \\

\bottomrule
\end{tabular}
\end{table}
{
As illustrated in Table~\ref{tab:transfer}, \sys achieves higher target-model scores on five of the six benchmarks and higher retention ratios on four of the six benchmarks. Compared with SkillOpt, \sys improves the average target-model score from 63.1\% to 67.8\% and the average retention ratio from 76.2\% to 81.3\%. The advantage remains after excluding ALFWorld, indicating that the improvement is not driven by a single benchmark; SearchQA is the only target-score exception.

These results show that the skills optimized by \sys transfer more effectively across agent backbones. A plausible explanation is that Optimization Memory consolidates recurring problems and evaluated solution attempts across iterations, encouraging the final skill to capture task-level procedures rather than source-model-specific wording. The transfer experiment evaluates the complete framework, however, and therefore does not isolate which component is responsible for the gain.
}

\subsection{Skill Quality Analysis}
\label{sec:skill-quality}

Beyond task performance, we evaluate the optimized skills with a frozen GPT-5.5 judge. The comparison covers ten benchmark slices, including six individual benchmarks and four DeepPlanning slices. Each skill is evaluated through three independent calls to the same judge model. The input does not reveal the method name. Each call assigns a score from 1 to 5 on six dimensions. Task Alignment and Non-Obvious Insight measure relevance. Constraint Handling and Decision Framework measure operational guidance. Cognitive Load and Inference Efficiency measure usability. For each dimension, we take the median of the three scores. The overall score for a skill is the mean of these six medians. We then average the overall scores across the ten benchmark slices.

\begin{figure}[t]
    \centering
    \includegraphics[width=0.99\linewidth]{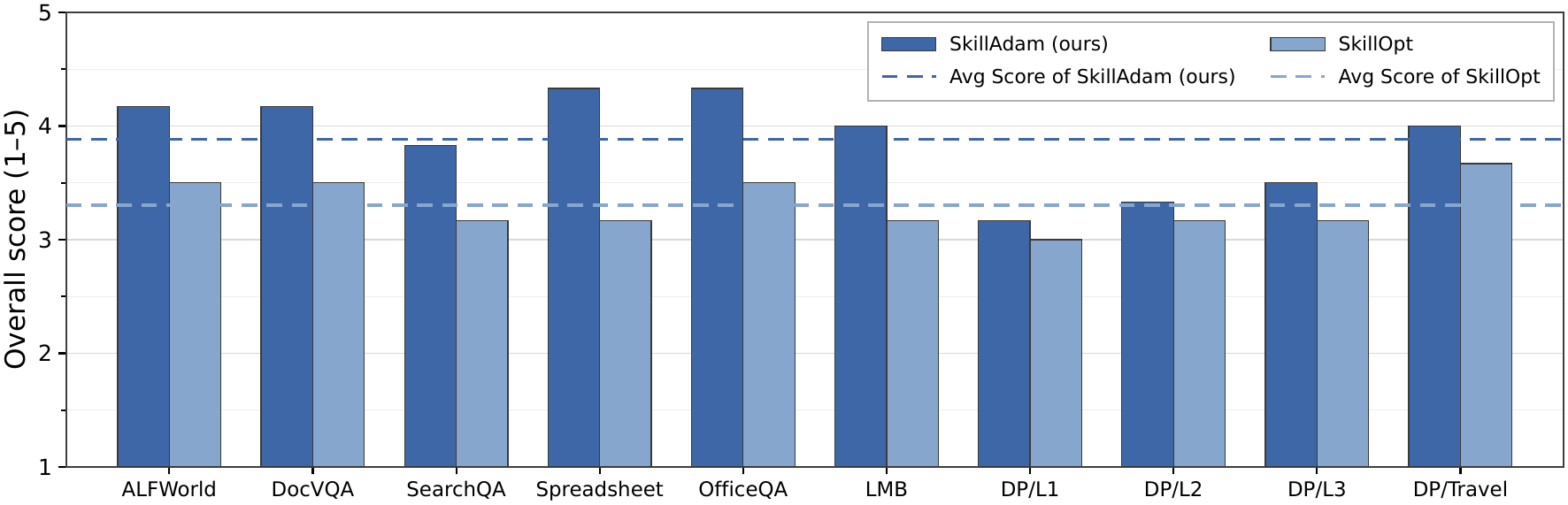}
    \caption{Overall skill-quality scores on a 1--5 scale across ten benchmark slices.
Each value averages the six dimension medians from three independent calls to
the same GPT-5.5 judge. Horizontal markers show the mean across slices.}
    \label{fig:judge}
\end{figure}

{
As illustrated in Figure~\ref{fig:judge}, \sys obtains a higher skill-quality score than SkillOpt on all ten benchmark slices, increasing the mean score from 3.30 to 3.88. Because every paired comparison favors \sys, the improvement is consistent across tasks rather than being driven by a small number of benchmarks.

This result complements the task-performance results in Section~\ref{sec:exp-main}. The benchmark metrics measure whether the agent completes the task, whereas the judge evaluates whether the skill provides relevant, operational, and usable guidance. The agreement between the two evaluations suggests that \sys improves not only downstream execution but also the quality of the skill document itself. This improvement is consistent with Optimization Memory preserving useful constraints and the Volatility-driven Edit Budget limiting uncontrolled revisions. However, the judge experiment evaluates the complete framework, and Figure~\ref{fig:judge} aggregates all six dimensions; it therefore isolates neither an individual component nor a single quality dimension.
}

\subsection{Training Cost}
\label{sec:exp-cost}

We compare optimization-phase API use for \sys and SkillOpt on the four DeepPlanning slices. Both methods use Claude Sonnet 4.5 and optimize over the same cases. The counts cover the optimization and iteration phase. They exclude initial-skill generation and final test evaluation. Unrelated smoke tests are also excluded. Token counts come from the raw counters returned by the API and do not represent a dollar cost. We report input tokens, output tokens, and API requests together with the resulting DeepPlanning performance.

\begin{table}[t]
\centering
\caption{Optimization-phase API use and DeepPlanning test performance. API-use deltas
are relative reductions. Performance deltas are absolute percentage-point
differences.}
\label{tab:cost}
\begin{tabular}{lccc}
\toprule
Metric & SkillOpt & \sys & Delta \\
\midrule
Input Tokens      & 222.1M & 72.5M & $-$67.3\% \\
Output Tokens     & 4.5M & 1.5M & $-$66.8\% \\
Total Tokens      & 226.6M & 74.0M & $-$67.3\% \\
API Requests      & 9,071 & 2,830 & $-$68.8\% \\
\midrule
\multicolumn{4}{l}{\textit{Resulting test performance (higher is better)}} \\
\midrule
DP-Shopping       & 41.7\% & \textbf{45.0\%} & $+3.3$ pp \\
DP-Travel         & 1.7\% & \textbf{11.7\%} & $+10.0$ pp \\
DP-Avg            & 21.7\% & \textbf{28.3\%} & $+6.7$ pp \\
\bottomrule
\end{tabular}
\end{table}

{
As illustrated in Table~\ref{tab:cost}, \sys reduces total token consumption by 67.3\% and API requests by 68.8\% compared with SkillOpt, while improving DP-Avg from 21.7\% to 28.3\%. Thus, the lower optimization cost is not achieved by sacrificing the quality of the final skill.

The two methods consume a similar number of tokens per request, and \sys is slightly higher on this measure. Therefore, the reduction does not come from shorter individual calls, but from requiring fewer optimization requests and modification attempts. This result is consistent with the two technical designs of \sys: Optimization Memory avoids repeatedly rediscovering previously identified failures, while the Volatility-driven Edit Budget reduces broad, weakly supported revisions. Measured by DP-Avg per million optimization tokens, \sys is approximately four times as efficient as SkillOpt.
}

\section{Analyses}
\label{sec:exp-dynamics}

\begin{figure}[t]
    \centering
    \includegraphics[width=0.8\linewidth]{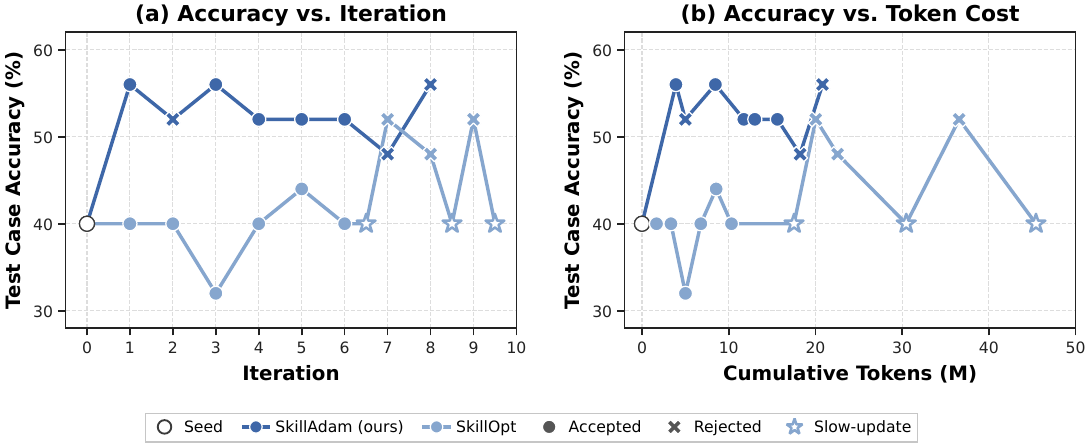}
    \caption{Optimization dynamics on DeepPlanning Shopping Level~1.
}
    \label{fig:dynamics}
\end{figure}

We analyze \sys and SkillOpt on DeepPlanning Shopping Level~1. Both methods use Claude Sonnet 4.5 and start from the same initial skill, which has 40\% test case accuracy. One iteration denotes one minibatch-level modification attempt. \sys compares the current skill with its proposed revision on the same sampled optimization cases and applies its multi-metric acceptance gate. SkillOpt follows its native selection and slow-update rules. The markers in Figure~\ref{fig:dynamics} therefore record the decision made by each method's own update protocol.

The figure reports the post-hoc test accuracy of every evaluated modification. Panel~(a) uses the minibatch-level iteration index, and Panel~(b) uses cumulative token consumption. Test performance is shown only for analysis and does not determine whether a modification is accepted. A modification with a high test score can still be rejected by the cases and metrics used in the corresponding update protocol.

\subsection{Stability}
\label{sec:exp-stability}

{
As illustrated in Figure~\ref{fig:dynamics}(a), \sys finds a strong revision in the first iteration, and all subsequent accepted skills remain above the initial performance. In contrast, SkillOpt's accepted regular updates fluctuate around or below its starting performance, while its later high-scoring candidates are not retained. This comparison shows that \sys produces a more stable accepted optimization path.

The result is consistent with the design of Optimization Memory. By retaining previously identified issues, their status, and the outcomes of prior solutions, \sys can incorporate new feedback without repeatedly overwriting useful corrections. The acceptance gate further prevents insufficiently supported revisions from replacing the current skill. The rejected \sys revision at iteration~8 nevertheless has a high post-hoc test score. This does not contradict the stability result because acceptance is determined only by sampled optimization cases and protected metrics, while the test score is used solely for retrospective analysis. Since the figure reports one run and memory operates together with the gate, it characterizes the overall optimization behavior rather than isolating the causal effect of memory alone.
}

\subsection{Efficiency}
\label{sec:exp-efficiency}

{
As illustrated in Figure~\ref{fig:dynamics}(b), \sys reaches an accepted skill with 56\% test accuracy after approximately 4M tokens, whereas SkillOpt first produces a comparably strong candidate after approximately 20M tokens and does not accept it. \sys therefore reaches and retains a stronger skill with about one fifth of the token cost in this run.

Combined with the similar token cost per request in Table~\ref{tab:cost}, the result shows that the efficiency gain comes from fewer unproductive modification attempts rather than cheaper individual calls. This behavior is consistent with the two core mechanisms: the Volatility-driven Edit Budget restricts broad edits when recent case-level effects are inconsistent, and Optimization Memory prevents repeated rediscovery of earlier failures. Together, they allow \sys to identify useful revisions with fewer iterations.
}

\section{Conclusion}

We introduced \textbf{\sys}, a gradient-inspired framework for iterative skill self-evolution that adapts the two principles behind Adam to the skill space. An \emph{optimization memory} accumulates structured records across iterations to provide stability, and a \emph{volatility-driven edit budget} scales each update to the reliability of the improvement evidence to provide adaptivity. Across seven benchmarks that span short-horizon and long-horizon agentic tasks, \sys achieves state-of-the-art performance at substantially lower training cost. Its skills also transfer better across models.

\bibliography{references}
\bibliographystyle{iclr2027_conference}

\end{document}